\documentclass[letterpaper, 10 pt, conference]{ieeeconf}  

\IEEEoverridecommandlockouts                              
\usepackage{graphics} 
\usepackage{epsfig} 
\usepackage{amsmath} 
\usepackage{amssymb}  

\usepackage{url}
\usepackage{verbatim} 
\usepackage{soul, color}
\usepackage{lmodern}
\usepackage{fancyhdr}
\usepackage[utf8]{inputenc}
\usepackage{fourier} 
\usepackage{array}
\usepackage{makecell}

\usepackage{algorithm}
\usepackage[noend]{algpseudocode}

\usepackage{multirow}

\makeatletter

\newcommand{\Rom}[1]{\expandafter\@slowromancap\romannumeral #1@}
\makeatother

\title{\LARGE \bf
Efficient Non-Compression Auto-Encoder for Driving Noise-based Road Surface Anomaly Detection
}

\author{YeongHyeon Park$^{*}$\thanks{$^*$Corresponding author: yeonghyeon@sk.com}, JongHee Jung 
\\ SK Planet Co., Ltd. \\
}

\begin{document}

\maketitle
\thispagestyle{plain}
\pagestyle{plain}

\begin{abstract}
Wet weather makes water film over the road and that film causes lower friction between tire and road surface. When a vehicle passes the low-friction road, the accident can occur up to 35\% higher frequency than a normal condition road. In order to prevent accidents as above, identifying the road condition in real-time is essential. Thus, we propose a convolutional auto-encoder-based anomaly detection model for taking both less computational resources and achieving higher anomaly detection performance. The proposed model adopts a non-compression method rather than a conventional bottleneck structured auto-encoder. As a result, the computational cost of the neural network is reduced up to 1 over 25 compared to the conventional models and the anomaly detection performance is improved by up to 7.72\%. Thus, we conclude the proposed model as a cutting-edge algorithm for real-time anomaly detection.

\end{abstract}


\section{Introduction}
\label{sec:introduction}
Comprehending the road surface status is highly helpful for preventing vehicle accidents. In prior research, Hall et al. have shown the wet-weather not only increases water film thickness of the road but also reduces friction coefficient between tire and surface~\cite{hall2009guide}. In a lower friction surface case, the vehicle can lose skid resistance and it can trigger the accident with high probability.

In another report, CM McGovern et al. have analyzed the road accidents occurred increased by $20\%$ and $35\%$ during wet weather than dry weather at Virginia and New York respectively~\cite{mcgovern2011state}. Paying attention in low-friction situations such as rainy cases is the best way to prevent accidents. However, drivers will be tired quickly in the above situation. 

\begin{figure}[h]
    \begin{center}
		\includegraphics[width=1.0\linewidth]{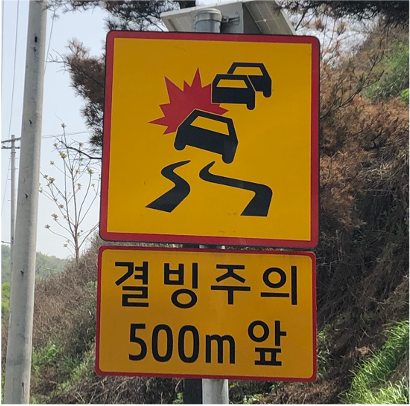}
	\end{center}
	\vspace*{-5mm}
	\caption{The warning signs that inform habitually icy sections. Although warning signs give a sense of alertness, the usefulness is limited because they do not indicate the current road surface condition.}
	\label{fig:sign}
\end{figure}

Moreover, installing signs to notify danger is widely used as shown Figure~\ref{fig:sign}, but it does not reflect the present road surface condition. Thus, if an anomaly detection function is provided as an advanced driver assistance system, it can be highly helpful to reduce both human exhaustion and road accidents simultaneously.

In prior research, there have been efforts to identify road surface types using computer vision~\cite{sy2008detection, raj2012vision, shen2014vision, nolte2018clf}. However, when using a vision sensor, it is difficult to properly define the road surface condition in situations such as blind spots or cloudy conditions. Also, when using the image to define the status of the road surface, the computational cost is larger than processing audio since the amount of information of the image collected within the same amount of time is greater than that of the sound~\cite{abdic2016clf, dogan2017svm, oh2018residual, park2018fared}. Thus, we adopt an audio sensor to construct an anomaly detection system.

Recently, many studies use generative neural networks as anomaly detection models~\cite{chauhan2015lstm, malhotra2015long, oh2018residual, park2018fared, wen2019time, jiang2019gan, zhang2019lvead, donahue2016bigan, thomas2017anogan, haowen2018vaekpi, wang2019advae, memarzadeh2020cvae, samet2018ganomaly, park2020hpgan}. The general premise of those anomaly detection models is that model, learned only normal state data, cannot generate abnormal state data well. Referring to the above, when the error between input and output generated from the anomaly detection model is higher, the state of input can be decided as abnormal. 

Among the prior research, there is a limitation in the qualitative evaluation when the anomaly detection model generates blurred output. One of the reasons is using variational bound and the other is parameter sharing between encoder and decoder~\cite{haowen2018vaekpi, wang2019advae, memarzadeh2020cvae, donahue2016bigan, thomas2017anogan}. The above limitation is not only solved by using feature matching and multiple hypotheses but also improved anomaly detection performance~\cite{samet2018ganomaly, park2020hpgan}. However, the limitation of those is the model becomes more complex.

For serving the system of road surface anomaly detection, numerous sensors and edge devices that correspond one-to-one for sensor, are required. Thus, we need to reduce the computational complexity to enable real-time processing in resource-constrained edge devices. The contributions for achieving the above goal are the following.

\begin{itemize}
   \item We propose novel neural network structure to reduce computational complexity.
   \item We compare and present the superiority of our model with others in quantitative and qualitative perspectives.
\end{itemize}

\section{Related work}
\label{sec:related_work}
In this section, we conduct the literature review of time-series data processing and structure of the neural networks for anomaly detection.

\subsection{Reflecting time-series causality}
\label{subsec:reflect_causality}
For handling time-series data such as audio data, reflecting causality from past to specific moment is one of the important things. Before using the deep learning-based technologies, some machine learning models such as the hidden Markov models or Kalman filter have adopted and used to reflect causality~\cite{ingmar2011hmm, ralaivola2005kalman}. However, the deep learning-based model outperformed the above conventional machine learning algorithm~\cite{deCruyenaere1992kalman, coskun2017kalman}. Moreover, the deep learning algorithms enable data-driven learning, that makes easy to solve the problems on the various field. For example, in our case, there is no need to define anomaly detection conditions by subdividing the differences in driving noise for each vehicle type such as car, truck or bus.

The advanced researches for processing time-series information, there are two types of the neural network are existed. One of them is a recurrent layer-based model and the other one is a convolutional layer-based model as known as recurrent neural networks (RNN) and convolutional neural networks (CNN) respectively~\cite{mikolov2010recurrent, graved2012lstm, chauhan2015lstm, malhotra2015long, potes2016heart, werbos1990bptt, ahmad2004bptt, chen2016gentle}. Refer that, long short-term memory (LSTM)~\cite{graved2012lstm, chauhan2015lstm, malhotra2015long} is one of the RNN structures.

In prior research RNN based model has achieved state-of-the-art performance at sound-based anomaly detection and fast adaptation to the changing environments~\cite{park2018fared}. The above model, fast adaptive RNN encoder-decoder (FARED), is a proper example of reflecting the causality of time-series information using RNN to solve the anomaly detection task. The FARED has a shallow layer, so training can be completed with a few epochs based on few parameters. However, if training with the same training iteration, the RNN structured model needs more time to update the parameters. The cause of the above is that RNN updates its parameter by backpropagation through the time algorithm~\cite{werbos1990bptt, ahmad2004bptt, chen2016gentle}.

Also, FARED can only reflect forward-way causality because it does not consider the bidirectional method for model construction~\cite{schuster1997birnn}. In RNN, the bidirectional method highly increases computational complexity and it also causes time consumption for training and inference procedure. Thus, we recommend that simple trial to improve anomaly detection performance should not be conducted via using the bidirectional method.

For easing the above limitation, CNN can be considered to construct the neural network architecture~\cite{oh2018residual, wen2019time, jiang2019gan, memarzadeh2020cvae, samet2018ganomaly, park2020hpgan}. In image processing, the convolutional filter slides on the input data and aggregates the spatial information for generating results. The other case in the above process, when the input data includes time series information, the convolutional filter aggregates and reflects the time information for generating output. Moreover, convolutional filter can reflect forward and backward causality without any settings such as considering bidirectional method. 

\subsection{Limitation of bottleneck structure}
\label{subsec:limitation_bottleneck}
The above studies, using generative CNN for time-series anomaly detection, commonly adopt bottleneck structure including encoder and decoder~\cite{oh2018residual, wen2019time, jiang2019gan, memarzadeh2020cvae, samet2018ganomaly, park2020hpgan}. The bottleneck structure compresses high-dimensional input to low-dimensional latent information via the encoder. Then, the decoder reconstructs the latent information as original high-dimensional information. 

For encoding and decoding the input data elaborately, the neural network should be constructed with sufficient capacity. Refer to the above, a bottleneck structured neural network needs much more layers to reconstruct from input to the output exquisitely which makes it difficult to construct lightweight architecture.

\section{Proposed approach}
\label{sec:proposed_approach}
We present the CNN-based anomaly detection model in this section. First of all, we describe the concept of road surface anomaly detection via vehicle driving noise. Sequentially after them, we present the architecture of the neural network and preprocessing method.

\subsection{Concept for anomaly detection}
\label{subsec:concept_anomaly_detection}
In view of the increased incidence of accidents on low-friction roads, we simply define the dry conditional surface as normal, and the wet condition as abnormal. We adopt the reconstruction error-based method using auto-encoder for anomaly detection as the same concept as earlier researches. Thus, we train the auto-encoder using normal samples only~\cite{chauhan2015lstm, malhotra2015long, oh2018residual, park2018fared, wen2019time, jiang2019gan, zhang2019lvead, donahue2016bigan, thomas2017anogan, haowen2018vaekpi, wang2019advae, memarzadeh2020cvae, samet2018ganomaly, park2020hpgan}.

\subsection{Non-compression auto-encoder}
\label{subsec:ncae}
In this research, we present a deep learning-based novel lightweight anomaly detection architecture. The proposed model is one of the convolutional auto-encoder but not a typical bottleneck architecture.

The complex and heavy architecture will consume more time for decisions and it causes informing delay. Note that delayed information does not useful to prevent or response for danger situation. Moreover, if the bottleneck structured model has sufficient capacity, it can probably reconstruct well even abnormal input. Because, bottelneck structured architecture generates new informations correspond to higher dimensions in decoding process~\cite{kingma2013auto, goodfellow2014generative, devries2017dataset, kim2021semiorthogonal}. In the end, it needs a lot of effort for tuning to get more lightweight architecture while only reconstructing the normal input successfully.

For minimizing the above efforts to find proper architecture, the non-bottleneck structure, named FARED, is proposed~\cite{park2018fared}. The FARED shows semiconductor assembly equipment anomaly detection performance similar to or higher with only 1 over 10 amount of the parameter than prior bottleneck structured CNN model. In addition, the FARED requires only a short time for training that enables quick response to target equipment changing.

\begin{figure}[h]
    \begin{center}
		\includegraphics[width=1.0\linewidth]{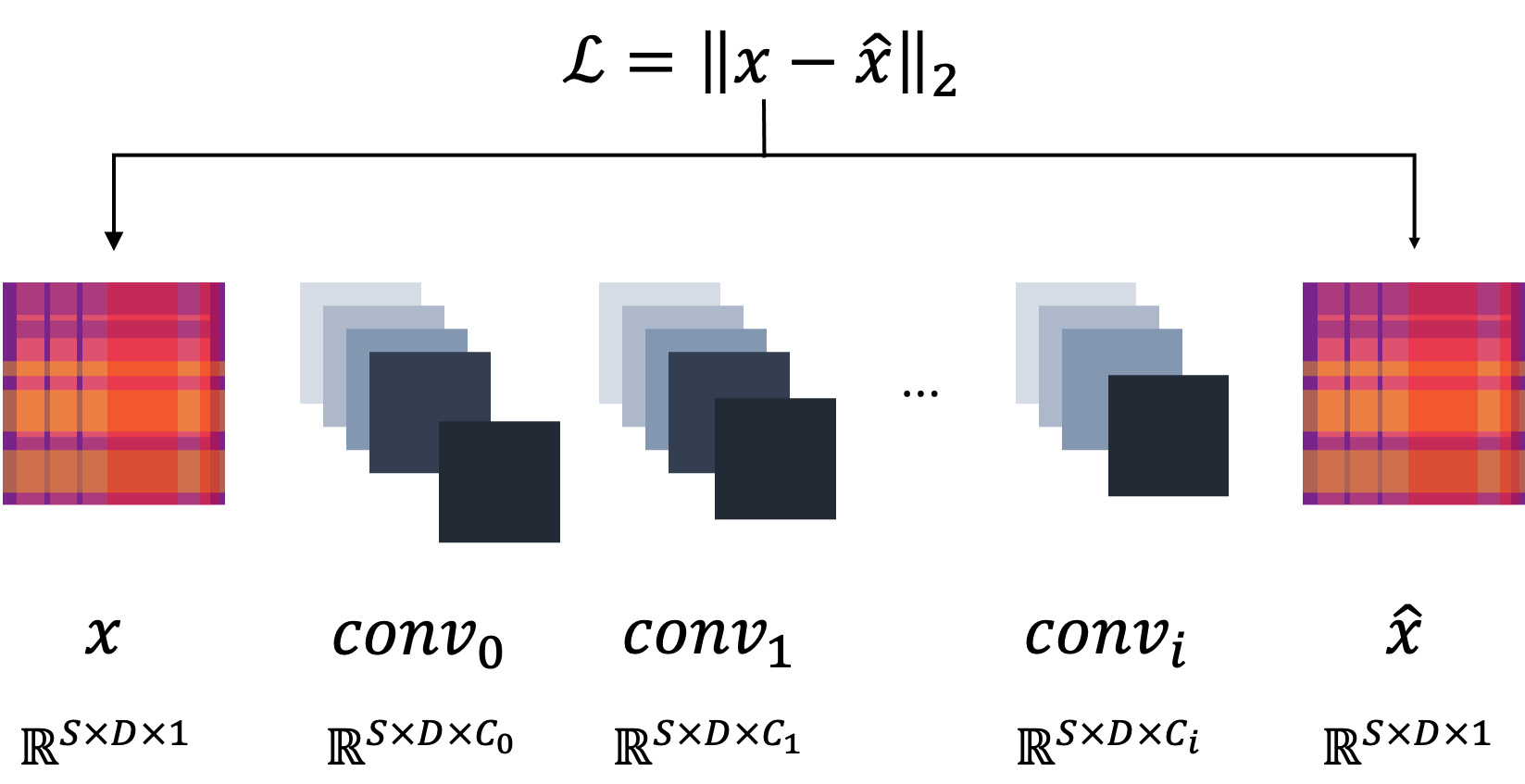}
	\end{center}
	\vspace*{-5mm}
	\caption{Architecture of the non-compressed auto-encoder (NCAE). NCAE performs only channel expansion and contraction operations through convolution operation without reduction of the time axis $S$ and the frequency axis $D$. Training is processed in a direction of minimizing the Euclidean distance of input $x$ and output $\hat{x}$.}
	\label{fig:ncae}
\end{figure}

We propose a CNN-based non-bottleneck structured auto-encoder referring to that non-bottleneck structure enables reducing the number of parameters with maintaining high anomaly detection performance. Also bidirectional causality can be reflected via the convolutional filter. We name the above neural network as non-compression auto-encoder (NCAE) because it does not compress the input information to the low-dimensional tensor. The architecture of NCAE is shown in Figure~\ref{fig:ncae}.

The NCAE has only transform and transition process from input to output, not the encoding and decoding process. According to the above method, there is no need to complicate the neural network architecture in order to perform the encoding and decoding elaborately. Also, the time consumption for training and test can be shortened as mentioned above.

Each convolutional layer is designed to remember the appropriate information at the filter for converting inputs to outputs. Thus, the NCAE leans to replicating the normal data only via combining each filter information. We use this property for anomaly detection. For example, when the abnormal data input to NCAE, the reconstruction error between input and ouput will be larger, because NCAE is designed to learn the reconstructing normal data only.

\subsection{Preprocessing}
\label{subsec:preprocessing}
Before feeding the input to the neural network, we conduct feature extraction as proposed in prior research~\cite{park2018fared}. The method, Mel-frequency cepstral coefficients (MFCC) extraction, have already achieved improved performance and computational efficiency for sound recognition~\cite{ahmad2004bptt, han2006speech, muda2010voice, zhang2016feature, foggia2016mfcc, socoro2017mfcc, almaadeed2018mfcc}. Thus, we adopt audio to MFCC converting algorithm for preprocessing.

We construct Mel-spectra for every 250ms time with 500ms length. First, we conduct a short-time Fourier transform (STFT) with 2048 for window length, and 512 for hopping. Then, we convert from spectra, the result of above, to the Mel-spectra with 128 dimensions of MFCC spectrum. The Mel-spectra are compressed to a single spectrum via time-averaging and we call it as MFCC vector. The MFCC vectors will stack over time. We use the above stacked vectors as input to the neural networks when the number of the stacked vectors is equal to $S$, i.e. $S$ sequence. These procedures are arranged as shown in Algorithm~\ref{algo:preprocessing}.
 
\begin{algorithm}
	\caption{Procedure of preprocessing}
	\begin{algorithmic}[1]
	    \State $F$: sampling rate
	    \State $W$: window length for Fourier transform
	    \State $H$: hopping length for Fourier transform
	    \State $D$: dimension of the MFCC vector
	    \State $s$: number of stacked MFCC vector
	    \State $S$: number of target stacks 
		\While{audio stream}
    		\State wait for 250ms 
    		\If{stream length $\geq$ 500ms} 
    		    \State split audio with 500ms: ${A}\in{\mathbb{R}^{(F/2)}}$
    		    \State construct Mel-spectra: ${M}_{spectra}\in{\mathbb{R}^{D\times{(F/2/W)}}}$
    		    \State make time-averaged MFCC vector:
    		    \State \hspace*{1.0em} ${M}_{spectrum}\in{\mathbb{R}^{D}}$
    		    \State stack the MFCC vector: ${V}\in{\mathbb{R}^{s\times{D}}}$
    	    \Else 
    	        \State continue
            \EndIf
    		\If{$s = S$} 
    		    \State feed stacked MFCC vector to neural network: 
    		    \State \hspace*{1.0em} $x$ same as ${V}\in{\mathbb{R}^{S\times{D}}}$
    		\EndIf
		\EndWhile
	\end{algorithmic}
	\label{algo:preprocessing}
\end{algorithm}

\section{Experiments}
\label{sec:experiments}
In this section, we present the dataset and experimental results for proving the validity of proposed model NCAE. 

\begin{figure}[h]
    \begin{center}
	    \includegraphics[width=1.0\linewidth]{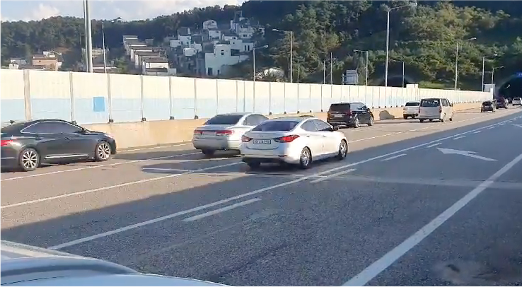} \\
	    \includegraphics[width=1.0\linewidth]{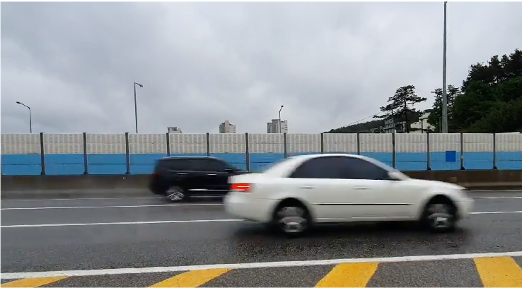}
	\end{center}
	\vspace*{-5mm}
	\caption{Environment for collecting the dataset. Audio is recorded at a 44.1 kHz sampling rate.}
	\label{fig:dataset}
\end{figure}

\subsection{Dataset}
\label{subsec:dataset}
Before experiments, we have collected the dataset of vehicle driving noise at Yongin-Seoul expressway with two weather condition, dry and wet. We collect the data to each condition as 20 minutes. Figure~\ref{fig:dataset} shows the data collecting environment.

We extract only driving event. Because vehicle driving noise-based road surface anomaly detection is impossible in a time section where no driving noise is not generated. We extract the noise section automatically via simple amplitude thresholding. Then, we manually discard non-driving noise such as horn sound, wind noise, or something. 

Developing the technology to automatically filter out those noises as mentioned above is not the scope of this study but we plan to cover them in the future work. The final dataset is summarized in Table~\ref{table:dataset}. We use 80\% of normal (dry) driving event for training and the others for the test.

\begin{table}[h]
    \centering
    \caption{Number of driving event and MFCC sequences for each weather condition \\}
    \begin{tabular}{lrr}
        \hline
        \textbf{} & \textbf{Driving event} & \textbf{Sequences} \\
        \hline
        Dry & 38 & 336 \\ 
        Wet & 27 & 243 \\ 
        \hline
    \end{tabular}
    \label{table:dataset}
\end{table}

\subsection{Training}
\label{subsec:training}
We need to conduct a training process first for measuring the anomaly detection performance of the proposed neural network architecture. Also, we need to compare our model to the previously published model.

After preprocessing, the input $X$ has three-axis $N$, $S$, and $D$ that means the number of sample $x$, length of sequences, and the number of MFCC feature. The neural network takes $X$ as input and generates the output $\hat{X}$ similar to $X$. Then, the neural network updates its parameters via backpropagation for minimizing Euclidean distance as shown in Equation~\ref{eq:target}.

\begin{equation}
    \label{eq:target}
    \mathcal{L} = ||X - \hat{X}||_{2} = \cfrac{1}{N}\sum_{n=1}^{N}\sqrt{\sum_{s=1,d=1}^{S, D}(x_{s,d} - \hat{x}_{s,d})}
\end{equation}

We adopt the Xavier initializer and Adam optimizer for parameter initialization and updating the parameters respectively~\cite{glorot2010xavier, kingma2017adam}. The training algorithm is described in Algorithm~\ref{algo:training}. 

\begin{algorithm}
	\caption{Training algorithm for anomaly detection model}
	\hspace*{1.0em}\textbf{Input} $X$: input \\
	\hspace*{2.0em} mini-batch constructed with stacked MFCC vector ${x}_{i}$ \\
    \hspace*{1.0em}\textbf{Output} $\hat{X}$: output mini-batch \\
    \hspace*{2.0em} corresponding to the input
	\begin{algorithmic}[1]
	    \State initialize network parameters by Xavier initializer~\cite{glorot2010xavier}
	    \While{until the loss converges}
    		\State get output $\hat{X}$ via neural network
    		\State compute Euclidean distance between $X$ and $\hat{X}$ 
    		\State \hspace*{1.0em} as Equation~\ref{eq:target}
    		\State update parameters by Adam optimizer~\cite{kingma2017adam}
		\EndWhile
	\end{algorithmic}
	\label{algo:training}
\end{algorithm}

\subsection{Comparison of computational costs}
\label{subsec:comparison_computation}
Prior to training, we construct four models besides NCAE for comparative experiments. The four models are FARED, auto-encoder (AE), variational auto-encoder (VAE), and HP-GAN. 

We have measured the number of the parameters and floating-point operations per second (FLOPs) of each model for each condition. The depth of the neural network module is limited to three because the goal of this study is not only to achieve higher anomaly detection performance but also to construct a lightweight neural network. The bottleneck structure has two neural network modules, an encoder and a decoder but the non-bottle neck has only a single module. Thus, the total number of layers for bottleneck and non-bottleneck are 6 and 3 respectively. In the bottleneck structured mode, the dimension of latent space is set as same as 128. The measurements of the above are summarized as shown in Table~\ref{table:consumption}.

\begin{table}[t]
    \centering
    \caption{Measurement of computing resource and time consumption. The MFLOPS means FLOPs in mega unit \\}
    
    \begin{tabular}{lccc}
        \hline
        \textbf{Model} & \textbf{Kernel size} & \textbf{num of params} & \textbf{MFLOPs} \\ 
        \hline
        FARED & - & \,\,\,\,594,432 & \,\,38.538 \\ 
        \hline
        \multirow{3}{*}{AE} & 3 & 2,726,144 & \,\,40.001 \\ 
            & 5 & 3,840,256 & \,\,65.233 \\ 
            & 7 & 4,954,368 & \,\,90.464 \\ 
        \hline
        \multirow{3}{*}{VAE} & 3 & 3,250,560 & \,\,41.050 \\ 
            & 5 & 4,364,672 & \,\,66.281 \\ 
            & 7 & 5,478,784 & \,\,91.513 \\ 
        \hline
        \multirow{3}{*}{HP-GAN} & 3 & 3,567,233 & 194.736 \\ 
            & 5 & 5,238,401 & 321.483 \\ 
            & 7 & 6,909,569 & 448.229 \\ 
        \hline
        \multirow{3}{*}{NCAE} & 3 & \,\,\,\,147,840 & \,\,\,\,8.863 \\ 
            & 5 & \,\,\,\,246,144 & \,\,14.761 \\ 
            & 7 & \,\,\,\,344,448 & \,\,20.659 \\ 
        \hline
    \end{tabular}
    
    \label{table:consumption}
\end{table}

Table~\ref{table:param_ratio} presents additional information that how much parameter and FLOPs NCAE has compared to other models are as a percentage. We use 3 for the kernel size to calculate the ratio as shown in Table~\ref{table:param_ratio} that can correspond to the minimum parameter of each model excluding the recurrent structured model. The number of parameters is reduced in maximum by 95.856\% compared to the previous model to the level of 4.144\%, and FLOPs are reduced by 95.449\% to the level corresponding to 4.551\%.

\begin{table}[h]
    \centering
    \caption{The cost of NCAE compared to others\\}
    
    \begin{tabular}{lcccc}
        \hline
            \textbf{} & \textbf{FARED} & \textbf{AE} & \textbf{VAE} & \textbf{HP-GAN} \\
        \hline
            Parameters & 24.870\% & \,\,\,5.423\% & \,\,\,4.548\% & \,\,\,4.144\% \\
            MFLOPs & 22.998\% & 22.156\% & 21.591\% & \,\,\,4.551\% \\
        \hline
    \end{tabular}
    
    \label{table:param_ratio}
\end{table}

\subsection{Hyperparameter tuning}
\label{subsec:hyperparameter_tuning}
Before comparing the several anomaly detection models in detail, we search the optimal hyperparameter for each anomaly detection model to set an equitable environment. We adopt the grid search method to find the best hyperparameter for each model~\cite{chicco2017ten}. We use the learning rate and kernel size as the common hyperparameters as shown in Table~\ref{table:hyperparameter}. 

However, the basic RNN structure, not combined with convolutional or some other layer only the learning rate is applied to the FARED as a hyperparameter. The bottleneck structured model has a latent vector and the dimension of them can be used as one of the hyperparameters but we only use 128-dimension for them as mentioned in Section~\ref{subsec:comparison_computation}.

\begin{table}[h]
    \centering
    \caption{Hyperparameters for the experiment. The symbol $z$ means latent vector. \\}
    
    \begin{tabular}{ll}
        \hline
        \textbf{Hyperparameter} & \textbf{Values} \\ 
        \hline
        Learning rate & 5e-3, 1e-3, 5e-4, 1e-4, 5e-5, and 1e-5 \\ 
        Kernel size & 3, 5, 7 \\ 
        Depth of layers & 3 for each module \\ 
        Dimension of $z$ & 128 in common \\ 
        \hline
    \end{tabular}
    
    \label{table:hyperparameter}
\end{table}

We have conducted training on neural networks and measured performance for each case. We use the area under the receiver characteristic curve (AUROC) as an anomaly detection performance indicator in this study~\cite{fawcett2006auroc}. The measured performance for each model with each training condition, called hyperparameter, is summarized in Table~\ref{table:performance_hyper_auroc}. 

\begin{table*}[h]
    \centering
    \caption{Measured AUROC for each model and hyperparameter. The bold text means the best performance for each model \\}
    
    \begin{tabular}{lccccccc}
        \hline
        \multirow{2}{*}{\textbf{Model}} & \multirow{2}{*}{\textbf{Kernel size}} & \multicolumn{6}{c}{\textbf{Learning rate}} \\
            &  & \textbf{5e-3} & \textbf{1e-3} & \textbf{5e-4} & \textbf{1e-4} & \textbf{5e-5} & \textbf{1e-5} \\ 
        \hline
        FARED & - & 0.92721 & \textbf{0.92738} & 0.92473 & 0.91927 & 0.90067 & 0.89188 \\ 
        \hline
        \multirow{3}{*}{AE} & 3 & 0.64268 & \textbf{0.96194} & 0.92866 & 0.91978 & 0.82207 & 0.70848 \\ 
            & 5 & 0.73881 & 0.87737 & 0.92686 & 0.93429 & 0.85347 & 0.67170 \\
            & 7 & 0.54659 & 0.68220 & 0.94129 & 0.90809 & 0.82412 & 0.67631 \\
        \hline
        \multirow{3}{*}{VAE} & 3 & 0.99420 & 0.99676 & \textbf{1.00000} & 0.96441 & 0.94718 & 0.96552 \\ 
            & 5 & 0.97781 & 0.98609 & 0.99846 & 0.98268 & 0.93301 & 0.97354 \\
            & 7 & 0.66270 & \textbf{1.00000} & 0.99428 & 0.95733 & 0.95579 & 0.99599 \\
        \hline
        \multirow{3}{*}{HP-GAN} & 3 & 0.69491 & 0.96834 & 0.97517 & 0.92695 & 0.95707 & 0.88377 \\ 
            & 5 & 0.76660 & 0.94231 & \textbf{0.99155} & 0.96237 & 0.97977 & 0.91884 \\
            & 7 & 0.68356 & 0.86918 & 0.96518 & 0.97022 & 0.90664 & 0.68894 \\
        \hline
        \multirow{3}{*}{NCAE} & 3 & 0.95417 & \textbf{1.00000} & \textbf{1.00000} & 0.92874 & 0.92388 & 0.91876 \\ 
            & 5 & 0.99684 & 0.97773 & 0.96091 & 0.93070 & 0.93361 & 0.91569 \\
            & 7 & \textbf{1.00000} & 0.98609 & 0.95366 & 0.93403 & 0.93685 & 0.93139 \\
        \hline
    \end{tabular}
    
    \label{table:performance_hyper_auroc}
\end{table*}

The proposed model, NCAE, achieves the highest AUROC of 1 three times, and the lowest performance is measured as 0.91569. The other model, VAE, achieves the AUROC of 1 two times but the lowest performance is 0.66270. Achieving high performance is essentially important, but it should be stable and easily achievable. In the above point of view, NCAE is effective and efficient than other models.

\begin{figure*}[h]
    \begin{center}
		\begin{tabular}{cc}
		    \includegraphics[width=0.45\linewidth]{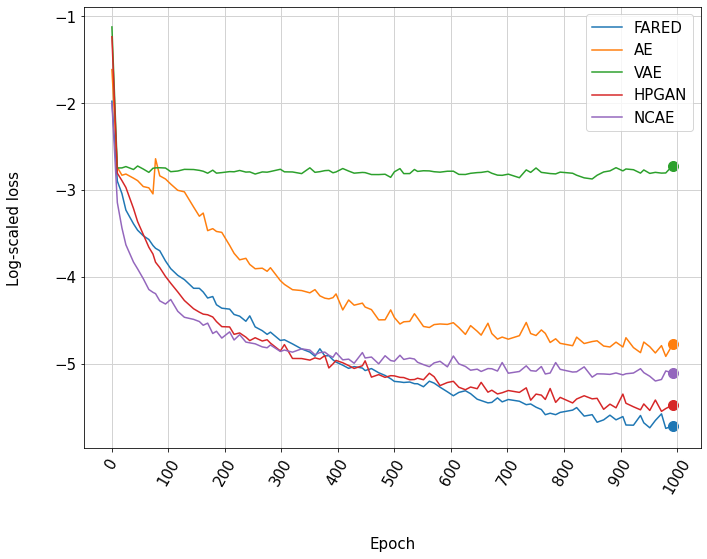} &
		    \includegraphics[width=0.45\linewidth]{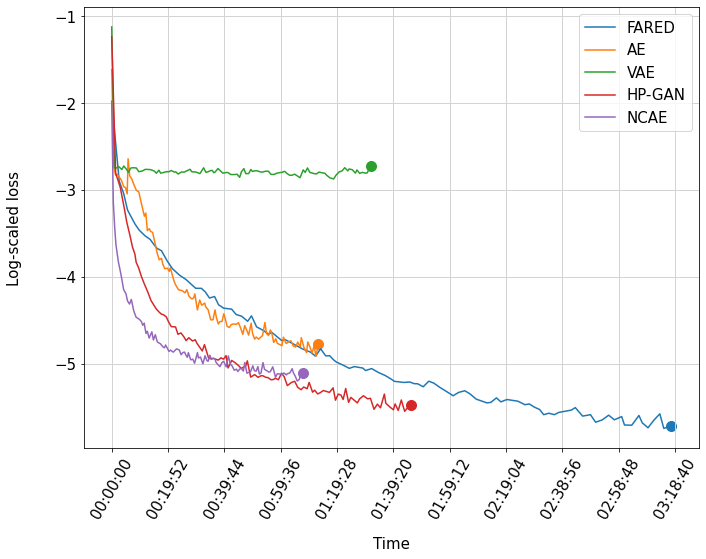}
		\end{tabular}
	\end{center}
	\vspace*{-5mm}
	\caption{Graph of loss changes at training process. Each graph shows loss per epoch and loss per time. For referring to the left figure, the loss of FARED is highly and continuously decreased. The loss of HP-GAN and NCAE is almost the similar scale and context as FARED. The training process ends in the order of NCAE, AE, VAE, HP-GAN, and FARED, which can be confirmed with the second graph.}
	\label{fig:loss_comparison}
\end{figure*}

Figure~\ref{fig:loss_comparison} shows loss convergence among the training process for each model. We can compare the degree of loss convergence in each model at the same number of training iterations via first graph of Figure~\ref{fig:loss_comparison}. We can also confirm which model finishes the training process earlier via the second graph.

\subsection{Verification with Monte Carlo estimation}
\label{subsec:montecarlo}
We conduct Monte Carlo estimation with five neural networks~\cite{kroese2014monte}. In Monte Carlo estimation, we evaluate the superiority including stability of the model through the mean and standard deviation of the performance. Each model is trained with the best-performing hyperparameters revealed by experiments in Section~\ref{subsec:hyperparameter_tuning}.

Table~\ref{table:performance_monte} shows the summarized AUROC and time consumption for each model. In anomaly detection performance evaluation, a high mean value of AUROC refers to high anomaly detection performance, and a low standard deviation refers to high learning stability. Accordingly, in order to construct a stable anomaly detection system, selecting a model with a high mean and low standard deviation is recommended. Referring to Table~\ref{table:performance_monte} and the above recommendation, we conclude the best model is NCAE, although other models also perform better. In the view point of time efficiency, NCAE is also better than others. Note that, time efficiency make ability to fast decision and explore more hyperparameters than others in the same amount of time.

\begin{table}[h]
    \centering
    \caption{Measured AUROC and time consumption with Monte Carlo estimation. The AUROC presented with minimum (Min), maximum (Max), mean, and standard deviation (SD) values. Time consumption shows only mean value. The best case is marked with bold text. \\}
    
    \begin{tabular}{lccccc}
        \hline
        \multirow{2}{*}{\textbf{Model}} & \multicolumn{1}{c}{\textbf{AUROC}} & \multicolumn{2}{c}{\textbf{Time Comsumption}} \\ 
        & \textbf{Mean+SD} & \textbf{Training} & \textbf{Inference} \\ 
        \hline
        FARED & 0.92819 $\pm$ 0.00302 & 03:13:54 & 0.06482  \\ 
        AE & 0.93428 $\pm$ 0.03057 & 01:13:22 & 0.01892 \\ 
        VAE & 0.98771 $\pm$ 0.02032 & 01:29:20 & 0.02486 \\ 
        HP-GAN & 0.97764 $\pm$ 0.01312 & 01:53:47 & 0.03273 \\ 
        NCAE & \textbf{0.99981 $\pm$ 0.00074} & \textbf{01:08:51} & \textbf{0.01749} \\ 
        \hline
    \end{tabular}
    
    \label{table:performance_monte}
\end{table}

\subsection{Qualitative analysis}
\label{subsec:qualitative}
\begin{figure*}[h]
    \centering
    \begin{tabular}{cc}
	    \multicolumn{2}{c}{\includegraphics[width=0.30\linewidth]{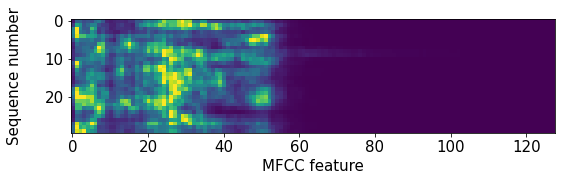}} \\ 
	    \includegraphics[width=0.30\linewidth]{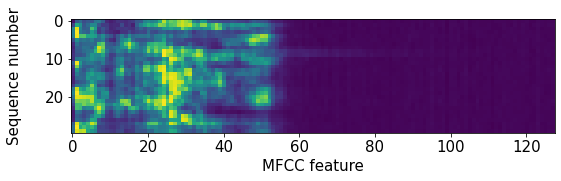} & \includegraphics[width=0.30\linewidth]{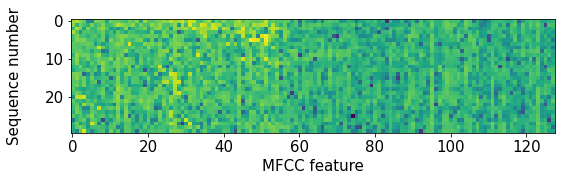} \\ 
	    \includegraphics[width=0.30\linewidth]{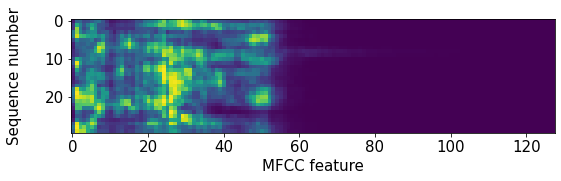} & \includegraphics[width=0.30\linewidth]{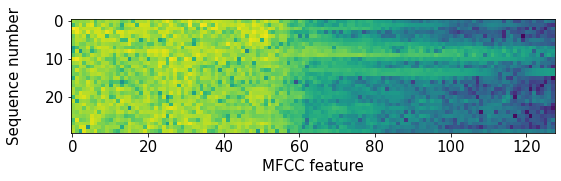} \\ 
	    \includegraphics[width=0.30\linewidth]{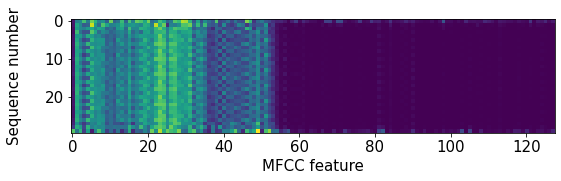} & \includegraphics[width=0.30\linewidth]{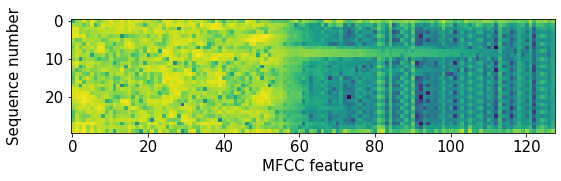} \\ 
	    \includegraphics[width=0.30\linewidth]{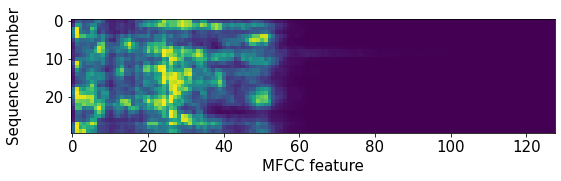} & \includegraphics[width=0.30\linewidth]{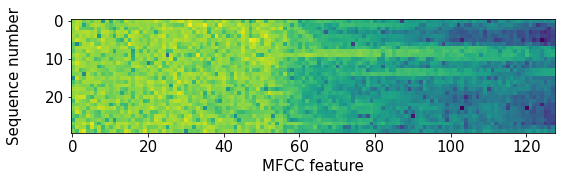} \\ 
	    \includegraphics[width=0.30\linewidth]{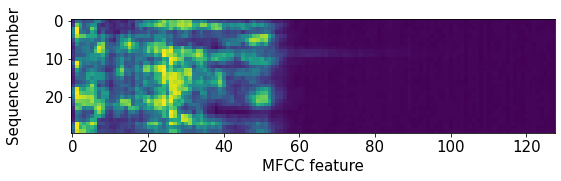} & \includegraphics[width=0.30\linewidth]{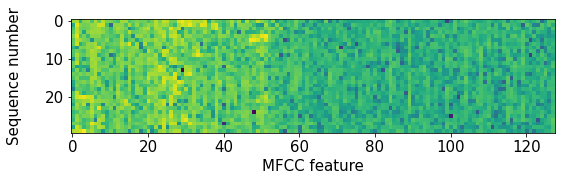} \\ 
	    \includegraphics[width=0.30\linewidth]{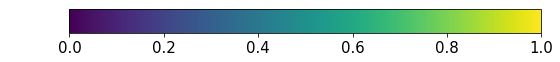} & \includegraphics[width=0.30\linewidth]{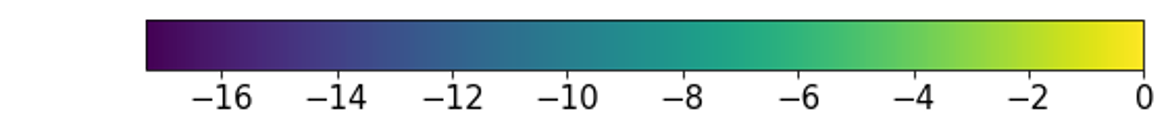} \\
	\end{tabular}
	\vspace*{-5mm}
	\caption{Result of reconstruction. The first figure shows input sequence. From the second row to the last row show the result of FARED, AE, VAE, HP-GAN, and NCAE sequentially. The left column shows the result of reconstruction, and the right column shows the log-scaled pixel-wise Euclidean distance between the input and reconstructed sequence.}
	\label{fig:qualitative}
\end{figure*}

We present the qualitative results for each model with Figure~\ref{fig:qualitative}. AE, VAE, and HP-GAN show lower restoration error in the high-frequency region, but in contrast, they do not restore the low-frequency region well. Even when an anomaly is revealed in a high-frequency region, it is important to restore the entire MFCC feature evenly in order to perform anomaly detection successfully. Moreover, VAE generates smoothed results over the sequence axis, same as the time axis, which makes it difficult to find the momentary anomaly. Accordingly, a model that restores uniformly and sharply as a whole region like NCAE is advantageous for anomaly detection.

\subsection{Discussion}
\label{subsec:discussion}

Sections~\ref{subsec:comparison_computation},~\ref{subsec:hyperparameter_tuning}, and~\ref{subsec:montecarlo} identify the advantages of using NCAE. Each of the above deals with computational cost, ease of hyperparameter tuning, and verified performance superiority through repeated experiments. When adopting NCAE as an anomaly detection model, enables minimizing resource consumption including time to update the system sustainably with maintaining high performance. Moreover, lower FLOPs of NCAE make it possible to execute without bottleneck effect even in limited computing environments such as edge devices.

For operating the anomaly detection system, the decision boundary can be set as Equation~\ref{eq:decision} referring to Tukey's fences basically~\cite{tukey1977exploratory}. The symbol $\mu$ and $\sigma$ are mean and standard deviation from the reconstruction error, as same as Equation~\ref{eq:target}, with the training set. 

\begin{equation}
    \label{eq:decision}
    \theta = \mu + (1.5 * \sigma)
\end{equation}

If the system operator wants to minimize missing out on anomalies, the threshold can be set to a lower value while taking a risk the error of decision the normal as abnormal and vise versa.

\section{Conclusion}
\label{sec:conclusion}
In this study, we propose a high-efficiency road surface anomaly detection model based on vehicle driving noise. The proposed model, NCAE, is composed of a non-compressional form instead of a bottleneck structure. NCAE does not need to construct both encoder and decoder for compression and reconstructing purposes. Moreover, there is no need to stack the layers deeply because there is no need to restore elaborately from the compressed lower-dimensional information to the original higher-dimension.

As a result, the amount of parameters is reduced to a minimum of 24.870\% and a maximum of 4.144\% level compared to previous models. Thus, the computational cost is also reduced from a minimum of 22.998\% to a maximum of 4.551\% compared to others. Through experiments, the anomaly detection performance of NCAE is improved by at least 1.23\% to up to 7.72\% compared to others. Also, the inference time is reduced at least 7.56\% and up to 73.02\%. Thus, we conclude NCAE is state-of-the-art architecture for sound-based road surface anomaly detection.

We will consider collecting more types of road anomalies, such as ice conditions, and we will conduct more anomaly detection experiments. On the other hand, we need to develop an automated method for discarding non-vehicle noise in future work. For the above method, developing an algorithm to classify each event can be considered. We can also consider researching a technique to reduce unnecessary noise or giving attention to important noise.

\section*{Acknowledgements}
We are grateful to all the members of SK Planet Co., Ltd., who have supported this research, data collection, providing equipment for the experiment.



\begin{thebibliography}{99}

\bibitem{hall2009guide}
Hall, JW and Smith, Kelly L and Titus-Glover, Leslie and Wambold, James C and Yager, Thomas J and Rado, Zoltan. Guide for pavement friction. \textit{Final Report for NCHRP Project 1} 2009; 43.

\bibitem{mcgovern2011state}
McGovern, Colleen M and Rusch, Peter F and Noyce, David A and others. State Practices to Reduce Wet Weather Skidding Crashes. \textit{United States. Federal Highway Administration. Office of Safety} 2011

\bibitem{mikolov2010recurrent}
Mikolov, Tom{\'a}{\v{s}} and Karafi{\'a}t, Martin and Burget, Luk{\'a}{\v{s}} and {\v{C}}ernock{\`y}, Jan and Khudanpur, Sanjeev, Recurrent neural network based language model. \textit{Eleventh annual conference of the international speech communication association} 2010; 2(3).

\bibitem{graved2012lstm}
Hochreiter, Sepp and Schmidhuber, J\"{u}rgen, Long Short-Term Memory. \textit{Neural Computation} 1997; 9(8): 1735-1780 

\bibitem{oh2018residual}
Oh, Dong Yul and Yun, Il Dong, Residual Error Based Anomaly Detection Using Auto-Encoder in SMD Machine Sound. \textit{Sensors} 2018; 18(5): 1308

\bibitem{park2018fared}
Park, YeongHyeon and Yun, Il Dong, Fast Adaptive RNN Encoder–Decoder for Anomaly Detection in SMD Assembly Machine. \textit{Sensors} 2018; 18(10): 3573

\bibitem{park2020hpgan}
Park, YeongHyeon and Park, Won Seok and Kim, Yeong Beom, Anomaly detection in particulate matter sensor using hypothesis pruning generative adversarial network. \textit{ETRI Journal} 2020

\bibitem{schuster1997birnn}
M. Schuster and K. K. Paliwal, Bidirectional recurrent neural networks. \textit{IEEE Transactions on Signal Processing} 1997; 45(11): 2673-2681

\bibitem{tukey1977exploratory}
Tukey, John W, Exploratory data analysis. \textit{Reading, Mass.} 1977; 2

\bibitem{kroese2014monte}
Kroese, Dirk P. and Brereton, Tim and Taimre, Thomas and Botev, Zdravko I., Why the Monte Carlo method is so important today. \textit{Wiley Interdisciplinary Reviews: Computational Statistics} 2014; 6(6); 386-392

\bibitem{hochreiter1997long}
Hochreiter, Sepp and Schmidhuber, J{\"u}rgen, Long short-term memory. \textit{Neural computation} 1997; 9(8); 1735-1780

\bibitem{chicco2017ten}
Chicco, Davide, Ten quick tips for machine learning in computational biology. \textit{BioData mining} 2017; 10(1); 1-17

\bibitem{kim2021semiorthogonal}
Jin-Hwa Kim and Do-Hyeong Kim and Saehoon Yi and Taehoon Lee, Semi-orthogonal Embedding for Efficient Unsupervised Anomaly Segmentation. \textit{arXiv preprint} 2021

\bibitem{ingmar2011hmm}
Ingmar Visser, Seven things to remember about hidden Markov models: A tutorial on Markovian models for time series. \textit{Journal of Mathematical Psychology} 2011; 55(6): 403-415

\bibitem{ralaivola2005kalman}
Ralaivola, L. and d'Alche-Buc, F., Time series filtering, smoothing and learning using the kernel Kalman filter. \textit{Proceedings. 2005 IEEE International Joint Conference on Neural Networks, 2005.} 2005; 3: 1449-1454

\bibitem{deCruyenaere1992kalman}
DeCruyenaere, J.P. and Hafez, H.M., A comparison between Kalman filters and recurrent neural networks. \textit{IJCNN International Joint Conference on Neural Networks} 1992; 4: 247-251

\bibitem{coskun2017kalman}
Coskun, Huseyin and Achilles, Felix and DiPietro, Robert and Navab, Nassir and Tombari, Federico, Long Short-Term Memory Kalman Filters: Recurrent Neural Estimators for Pose Regularization. \textit{Proceedings of the IEEE International Conference on Computer Vision (ICCV)} 2017

\bibitem{potes2016heart}
Potes, Cristhian and Parvaneh, Saman and Rahman, Asif and Conroy, Bryan, Ensemble of feature-based and deep learning-based classifiers for detection of abnormal heart sounds. \textit{2016 Computing in Cardiology Conference (CinC)} 2016; 621-624

\bibitem{han2006speech}
Wei Han and Cheong-Fat Chan and Chiu-Sing Choy and Kong-Pang Pun, An efficient MFCC extraction method in speech recognition. \textit{2006 IEEE International Symposium on Circuits and Systems} 2006; 4

\bibitem{muda2010voice}
Lindasalwa Muda and Mumtaj Begam and I. Elamvazuthi, Voice Recognition Algorithms using Mel Frequency Cepstral Coefficient (MFCC) and Dynamic Time Warping (DTW) Techniques. \textit{arXiv preprint} 2010

\bibitem{wen2019time}
Tailai Wen and Roy Keyes, Time Series Anomaly Detection Using Convolutional Neural Networks and Transfer Learning. \textit{arXiv preprint} 2019

\bibitem{jiang2019gan}
Jiang, Wenqian and Hong, Yang and Zhou, Beitong and He, Xin and Cheng, Cheng, A GAN-Based Anomaly Detection Approach for Imbalanced Industrial Time Series. \textit{IEEE Access} 2019; 7: 143608-143619

\bibitem{memarzadeh2020cvae}
Memarzadeh, Milad and Matthews, Bryan and Avrekh, Ilya, Unsupervised Anomaly Detection in Flight Data Using Convolutional Variational Auto-Encoder. \textit{Aerospace} 2020; 7(8)

\bibitem{kingma2017adam}
Diederik P. Kingma and Jimmy Ba, Adam: A Method for Stochastic Optimization. \textit{arXiv preprint} 2017

\bibitem{glorot2010xavier}
Glorot, Xavier and Bengio, Yoshua, Understanding the difficulty of training deep feedforward neural networks. \textit{Proceedings of the Thirteenth International Conference on Artificial Intelligence and Statistics} 2010; 9: 249-356

\bibitem{kingma2013auto}
Kingma, Diederik P and Welling, Max, Auto-encoding variational bayes. \textit{arXiv preprint} 2013

\bibitem{goodfellow2014generative}
Goodfellow, Ian J and Pouget-Abadie, Jean and Mirza, Mehdi and Xu, Bing and Warde-Farley, David and Ozair, Sherjil and Courville, Aaron and Bengio, Yoshua, Generative adversarial networks. \textit{arXiv preprint} 2014

\bibitem{devries2017dataset}
Terrance DeVries and Graham W. Taylor, Dataset Augmentation in Feature Space. \textit{arXiv preprint} 2017

\bibitem{fawcett2006auroc}
Tom Fawcett, An introduction to ROC analysis. \textit{Pattern Recognition Letters} 2006; 27(8): 861-874

\bibitem{chen2016gentle}
Chen, Gang, A gentle tutorial of recurrent neural network with error backpropagation. \textit{arXiv preprint} 2016

\bibitem{ahmad2004bptt}
Ahmad, A.M. and Ismail, S. and Samaon, D.F., Recurrent neural network with backpropagation through time for speech recognition. \textit{IEEE International Symposium on Communications and Information Technology, 2004. ISCIT 2004.} 2004; 1: 98-102

\bibitem{werbos1990bptt}
Werbos, P.J., Backpropagation through time: what it does and how to do it. \textit{Proceedings of the IEEE.} 1990; 78(10): 1550-1560

\bibitem{chauhan2015lstm}
Chauhan, Sucheta and Vig, Lovekesh, Anomaly detection in ECG time signals via deep long short-term memory networks. \textit{2015 IEEE International Conference on Data Science and Advanced Analytics (DSAA)} 2015; 1-7

\bibitem{malhotra2015long}
Malhotra, Pankaj and Vig, Lovekesh and Shroff, Gautam and Agarwal, Puneet, Long short term memory networks for anomaly detection in time series. \textit{Presses universitaires de Louvain} 2015; 89: 89-94

\bibitem{sy2008detection}
Sy, N. T. and Avila, M. and Begot, S. and Bardet, J. C., Detection of defects in road surface by a vision system. \textit{MELECON 2008 - The 14th IEEE Mediterranean Electrotechnical Conference} 2008; 847-851

\bibitem{raj2012vision}
Raj, Arjun and Krishna, Dilip and Hari Priya, R. and Shantanu, Kumar and Niranjani Devi, S., Vision based road surface detection for automotive systems. \textit{2012 International Conference on Applied Electronics} 2012; 223-228

\bibitem{shen2014vision}
Shen, Truman and Schamp, Gregory and Haddad, Mario, Stereo vision based road surface preview. \textit{17th International IEEE Conference on Intelligent Transportation Systems (ITSC)} 2014; 1843-1849

\bibitem{nolte2018clf}
Nolte, Marcus and Kister, Nikita and Maurer, Markus, Assessment of Deep Convolutional Neural Networks for Road Surface Classification. \textit{2018 21st International Conference on Intelligent Transportation Systems (ITSC)} 2018; 381-386

\bibitem{foggia2016mfcc}
Foggia, Pasquale and Petkov, Nicolai and Saggese, Alessia and Strisciuglio, Nicola and Vento, Mario, Audio Surveillance of Roads: A System for Detecting Anomalous Sounds. \textit{IEEE Transactions on Intelligent Transportation Systems} 2016; 17(1): 279-288

\bibitem{dogan2017svm}
Do\u{g}an, Da\u{g}han, Road-types classification using audio signal processing and SVM method. \textit{2017 25th Signal Processing and Communications Applications Conference (SIU)} 2017; 1-4

\bibitem{almaadeed2018mfcc}
Almaadeed, Noor and Asim, Muhammad and Al-Maadeed, Somaya and Bouridane, Ahmed and Beghdadi, Azeddine, Automatic Detection and Classification of Audio Events for Road Surveillance Applications. \textit{Sensors} 2018; 18(6): 1858

\bibitem{zhang2016feature}
Zhang, Lanyue and Wu, Di and Han, Xue and Zhu, Zhongrui, Feature extraction of underwater target signal using mel frequency cepstrum coefficients based on acoustic vector sensor. \textit{Hindawi} 2016

\bibitem{socoro2017mfcc}
Socor\'{o}, Joan Claudi and Al\'{i}as, Francesc and Alsina-Pag\`{e}s, Rosa Ma, An Anomalous Noise Events Detector for Dynamic Road Traffic Noise Mapping in Real-Life Urban and Suburban Environments. \textit{Sensors} 2017; 17(10)

\bibitem{abdic2016clf}
Abdi\`{c}, Irman and Fridman, Lex and Brown, Daniel E. and Angell, William and Reimer, Bryan and Marchi, Erik and Schuller, Björn, Detecting road surface wetness from audio: A deep learning approach. \textit{2016 23rd International Conference on Pattern Recognition (ICPR)} 2016; 3458-3463

\bibitem{haowen2018vaekpi}
Xu, Haowen and Chen, Wenxiao and Zhao, Nengwen and Li, Zeyan and Bu, Jiahao and Li, Zhihan and Liu, Ying and Zhao, Youjian and Pei, Dan and Feng, Yang and Chen, Jie and Wang, Zhaogang and Qiao, Honglin, Unsupervised Anomaly Detection via Variational Auto-Encoder for Seasonal KPIs in Web Applications. \textit{Proceedings of the 2018 World Wide Web Conference} 2018; WWW '18(10): 187-196

\bibitem{thomas2017anogan}
Schlegl, Thomas, Seeb{\"o}ck, Philipp ,Waldstein, Sebastian M. ,Schmidt-Erfurth, Ursula ,Langs, Georg, Unsupervised Anomaly Detection with Generative Adversarial Networks to Guide Marker Discovery. \textit{Information Processing in Medical Imaging} 2017; 146-157

\bibitem{donahue2016bigan}
Jeff Donahue, Philipp Kr{\"{a}}henb{\"{u}}hl, Trevor Darrell, Adversarial Feature Learning. \textit{arXiv preprint} 2016

\bibitem{samet2018ganomaly}
Akcay, Samet and Atapour-Abarghouei, Amir and Breckon, Toby P., GANomaly: Semi-supervised Anomaly Detection via Adversarial Training. \textit{Computer Vision - ACCV 2018} 2019

\bibitem{wang2019advae}
Xuhong Wang and, Ying Du and, Shijie Lin and, Ping Cui and, Yupu Yang, Self-adversarial Variational Autoencoder with Gaussian Anomaly Prior Distribution for Anomaly Detection. \textit{arXiv preprint} 2019

\bibitem{zhang2019lvead}
Chunkai Zhang, Yingyang Chen, Time Series Anomaly Detection with Variational Autoencoders. \textit{arXiv preprint} 2019

\end{thebibliography}

\end{document}